\documentclass[10pt,twocolumn,letterpaper]{article}

\usepackage{cvpr}
\usepackage{times}
\usepackage{epsfig}
\usepackage{graphicx}
\usepackage{amsmath}
\usepackage{amssymb}

\usepackage{multirow}

\usepackage[pagebackref=true,breaklinks=true,letterpaper=true,colorlinks,bookmarks=false]{hyperref}

\cvprfinalcopy 

\def\cvprPaperID{****} 

\ifcvprfinal\fi
\begin{document}

\title{Interpretable Convolutional Neural Networks}

\author{Quanshi Zhang, Ying Nian Wu, and Song-Chun Zhu\\
University of California, Los Angeles
}

\maketitle

\begin{abstract}
This paper proposes a method to modify traditional convolutional neural networks (CNNs) into interpretable CNNs, in order to clarify knowledge representations in high conv-layers of CNNs. In an interpretable CNN, each filter in a high conv-layer represents a specific object part. Our interpretable CNNs use the same training data as ordinary CNNs without a need for additional annotations of object parts or textures for supervision. The interpretable CNN automatically assigns each filter in a high conv-layer with an object part during the learning process. We can apply our method to different types of CNNs with various structures. The explicit knowledge representation in an interpretable CNN can help people understand logic inside a CNN, \emph{i.e.} what patterns are memorized by the CNN for prediction. Experiments have shown that filters in an interpretable CNN are more semantically meaningful than those in traditional CNNs.\footnote[1]{The code is available at \url{https://github.com/zqs1022/interpretableCNN}}.
\end{abstract}


\section{Introduction}

Convolutional neural networks (CNNs)~\cite{CNN,CNNImageNet,ResNet} have achieved superior performance in many visual tasks, such as object classification and detection. As discussed in Bau~\emph{et al.}~\cite{Interpretability}, besides the discrimination power, model interpretability is another crucial issue for neural networks. However, the interpretability is always an Achilles' heel of CNNs, and has presented considerable challenges for decades.

In this paper, we focus on a new problem, \emph{i.e.} \textit{without any additional human supervision, can we modify a CNN to obtain interpretable knowledge representations in its conv-layers?} We expect the CNN has a certain introspection of its representations during the end-to-end learning process, so that the CNN can regularize its representations to ensure high interpretability. Our learning for high interpretability is different from conventional off-line visualization~\cite{CNNVisualization_1,CNNVisualization_2,CNNVisualization_3,FeaVisual,visualCNN_grad,visualCNN_grad_2} and diagnosis~\cite{Interpretability,CNNInfluence,banditUnknown,trust} of pre-trained CNN representations.

\begin{figure}[t]
\centering
\includegraphics[width=0.99\linewidth]{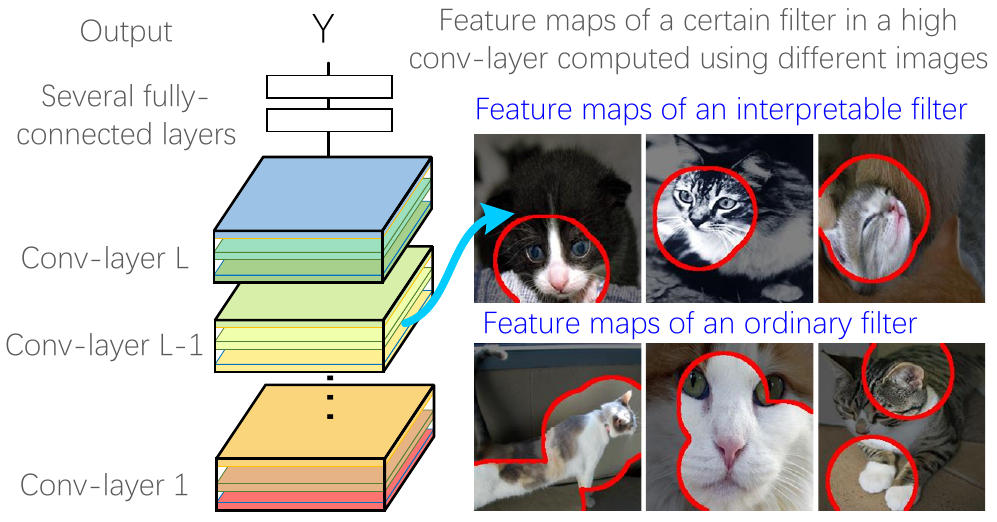}
\caption{Comparison of a filter's feature maps in an interpretable CNN and those in a traditional CNN.}
\label{fig:top}
\vspace{-8pt}
\end{figure}

Bau~\emph{et al.}~\cite{Interpretability} defined six kinds of semantics in CNNs, \emph{i.e.} \textit{objects}, \textit{parts}, \textit{scenes}, \textit{textures}, \textit{materials}, and \textit{colors}. In fact, we can roughly consider the first two semantics as object-part patterns with specific shapes, and summarize the last four semantics as texture patterns without clear contours. Moreover, filters in low conv-layers usually describe simple textures, whereas filters in high conv-layers are more likely to represent object parts.

Therefore, in this study, we aim to train each filter in a high conv-layer to represent an object part. Fig.~\ref{fig:top} shows the difference between a traditional CNN and our interpretable CNN. In a traditional CNN, a high-layer filter may describe a mixture of patterns, \emph{i.e.} the filter may be activated by both the head part and the leg part of a cat. Such complex representations in high conv-layers significantly decrease the network interpretability. In contrast, the filter in our interpretable CNN is activated by a certain part. In this way, we can explicitly identify which object parts are memorized in the CNN for classification without ambiguity. The goal of this study can be summarized as follows.
\begin{itemize}
\vspace{-5pt}
\item We propose to slightly revise a CNN to improve its interpretability, which can be broadly applied to CNNs with different structures.
\vspace{-5pt}
\item We do \textbf{not} need any annotations of object parts or textures for supervision. Instead, our method automatically pushes the representation of each filter towards an object part.
\vspace{-5pt}
\item The interpretable CNN does not change the loss function on the top layer and uses the same training samples as the original CNN.
\vspace{-5pt}
\item As an exploratory research, the design for interpretability may decrease the discrimination power a bit, but we hope to limit such a decrease within a small range.
\end{itemize}

\textbf{Methods:} Given a high conv-layer in a CNN, we propose a simple yet effective loss for each filter in the conv-layer to push the filter towards the representation of an object part. As shown in Fig.~\ref{fig:net}, we add a loss for the output feature map of each filter. The loss encourages a low entropy of inter-category activations and a low entropy of spatial distributions of neural activations. \emph{I.e.} each filter must encode a distinct object part that is exclusively contained by a single object category, and the filter must be activated by a single part of the object, rather than repetitively appear on different object regions. For example, the left eye and the right eye may be represented using two different part filters, because contexts of the two eyes are symmetric, but not the same. Here, we assume that repetitive shapes on various regions are more prone to describe low-level textures (\emph{e.g.} colors and edges), instead of high-level parts.

\textbf{The value of network interpretability:} The clear semantics in high conv-layers is of great importance when we need human beings to trust a network's prediction. In spite of the high accuracy of neural networks, human beings usually cannot fully trust a network, unless it can explain its logic for decisions, \emph{i.e.} what patterns are memorized for prediction. Given an image, current studies for network diagnosis~\cite{visualCNN_grad,visualCNN_grad_2,trust} localize image regions that contribute most to network predictions at the pixel level. In this study, we expect the CNN to explain its logic at the object-part level. Given an interpretable CNN, we can explicitly show the distribution of object parts that are memorized by the CNN for object classification.

\textbf{Contributions:} In this paper, we focus on a new task, \emph{i.e.} end-to-end learning a CNN whose representations in high conv-layers are interpretable. We propose a simple yet effective method to modify different types of CNNs into interpretable CNNs without any additional annotations of object parts or textures for supervision. Experiments show that our approach has significantly improved the object-part interpretability of CNNs.

\section{Related work}

The interpretability and the discrimination power are two important properties of a model~\cite{Interpretability}. In recent years, different methods are developed to explore the semantics hidden inside a CNN. Many statistical methods~\cite{CNNAnalysis_1,CNNAnalysis_2,CNNVisualization_5} have been proposed to analyze CNN features.

\textbf{Network visualization:} Visualization of filters in a CNN is the most direct way of exploring the pattern hidden inside a neural unit. \cite{CNNVisualization_1,CNNVisualization_2,CNNVisualization_3} showed the appearance that maximized the score of a given unit. up-convolutional nets~\cite{FeaVisual} were used to invert CNN feature maps to images.

\textbf{Pattern retrieval:}{\verb| |} Some studies go beyond passive visualization and actively retrieve certain units from CNNs for different applications. Like the extraction of mid-level features~\cite{MiddleLevel} from images, pattern retrieval mainly learns mid-level representations from conv-layers. Zhou~\emph{et al.}~\cite{CNNSemanticDeep,CNNSemanticDeep2} selected units from feature maps to describe ``scenes''. Simon~\emph{et al.} discovered objects from feature maps of unlabeled images~\cite{ObjectDiscoveryCNN_2}, and selected a certain filter to describe each semantic part in a supervised fashion~\cite{CNNSemanticPart}. \cite{CNNAoG} extracted certain neural units from a filter's feature map to describe an object part in a weakly-supervised manner. \cite{interpretVQA_grad} used a gradient-based method to interpret visual question-answering models. Studies of \cite{explainableFeature,explainableFeature2,explainableFeature3,explainableFeature4} selected neural units with specific meanings from CNNs for various applications.


\textbf{Model diagnosis: }{\verb| |} Many methods have been developed to diagnose representations of a black-box model. The LIME method proposed by Ribeiro~\emph{et al.}~\cite{trust}, influence functions~\cite{CNNInfluence} and gradient-based visualization methods~\cite{visualCNN_grad,visualCNN_grad_2} and \cite{ExplainingArea} extracted image regions that were responsible for each network output, in order to interpret network representations. These methods require people to manually check image regions accountable for the label prediction for each testing image. \cite{CNNDiagnosis} extracted relationships between representations of various categories from a CNN. Lakkaraju~\emph{et al.}~\cite{banditUnknown} and Zhang~\emph{et al.}~\cite{DeepQA} explored unknown knowledge of CNNs via active annotations and active question-answering. In contrast, given an interpretable CNN, people can directly identify object parts (filters) that are used for decisions during the inference procedure.


\textbf{Learning a better representation:} Unlike the diagnosis and/or visualization of pre-trained CNNs, some approaches are developed to learn more meaningful representations. \cite{rightReason} required people to label dimensions of the input that were related to each output, in order to learn a better model. Hu~\emph{et al.}~\cite{LogicRuleNetwork} designed some logic rules for network outputs, and used these rules to regularize the learning process. Stone~\emph{et al.}~\cite{CNNCompositionality} learned CNN representations with better object compositionality, but they did not obtain explicit part-level or texture-level semantics. Sabour~\emph{et al.}~\cite{capsule} proposed a capsule model, which used a dynamic routing mechanism to parse the entire object into a parsing tree of capsules, and each capsule may encode a specific meaning. In this study, we invent a generic loss to regularize the representation of a filter to improve its interpretability. We can analyze the interpretable CNN from the perspective of information bottleneck~\cite{InformationBottleneck} as follows. 1) Our interpretable filters selectively model the most distinct parts of each category to minimize the conditional entropy of the final classification given feature maps of a conv-layer. 2) Each filter represents a single part of an object, which maximizes the mutual information between the input image and middle-layer feature maps (\emph{i.e.} ``forgetting'' as much irrelevant information as possible).

\section{Algorithm}

Given a target conv-layer of a CNN, we expect each filter in the conv-layer to be activated by a certain object part of a certain category, and keep inactivated on images of other categories. Let {\small${\bf I}$} denote a set of training images, where {\small${\bf I}_{c}\subset{\bf I}$} represents the subset that belongs to category $c$, ({\small$c=1,2,\ldots,C$}). Theoretically, we can use different types of losses to learn CNNs for multi-class classification, single-class classification (\emph{i.e.} {\small$c=1$} for images of a category and {\small$c=2$} for random images), and other tasks.

\begin{figure}[t]
\centering
\includegraphics[width=0.99\linewidth]{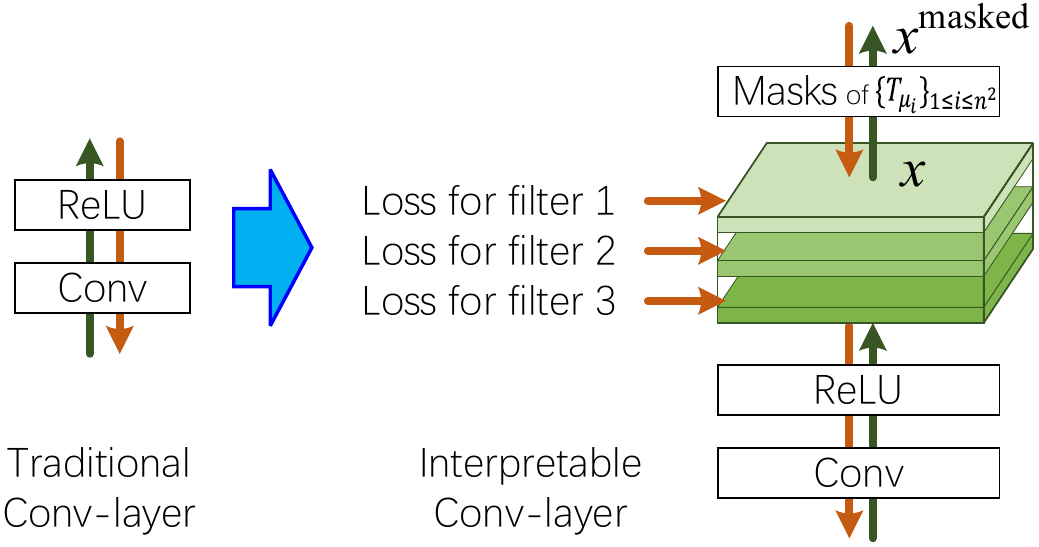}
\caption{Structures of an ordinary conv-layer and an interpretable conv-layer. Green and red lines indicate the forward and backward propagations, respectively.}
\label{fig:net}
\vspace{-8pt}
\end{figure}

\begin{figure}[t]
\centering
\includegraphics[width=0.99\linewidth]{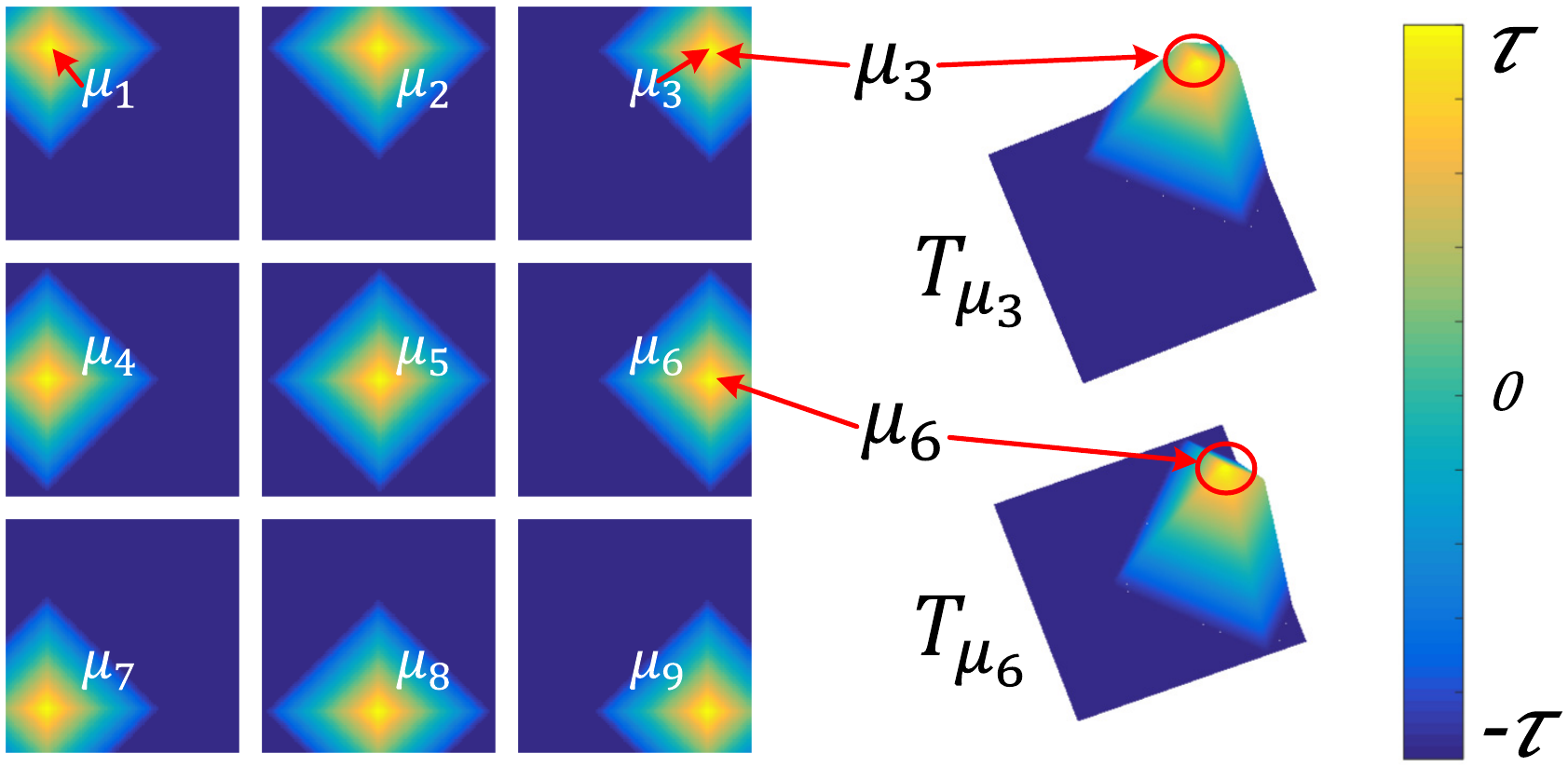}
\caption{Templates of {\small$T_{\mu_{i}}$}. In fact, the algorithm also supports a round template based on the L-2 norm distance. Here, we use the L-1 norm distance instead to speed up the computation.}
\label{fig:template}
\vspace{-8pt}
\end{figure}

Fig.~\ref{fig:net} shows the structure of our interpretable conv-layer. In the following paragraphs, we focus on the learning of a single filter $f$ in the target conv-layer. We add a loss to the feature map $x$ of the filter $f$ after the ReLu operation. The feature map $x$ is an {\small$n\times n$} matrix, {\small$x_{ij}\geq0$}. Because $f$'s corresponding object part may appear at different locations in different images, we design $n^2$ templates for $f$ {\small$T_{\mu_1},T_{\mu_2},\ldots,T_{\mu_{n^2}}\}$}. As shown in Fig.~\ref{fig:template}, each template {\small$T_{\mu_{i}}$} is also an {\small$n\times n$} matrix, and it describes the ideal distribution of activations for the feature map $x$ when the target part mainly triggers the $i$-th unit in $x$.

\textbf{During the forward propagation}, given each input image $I$, the CNN selects a specific template {\small$T_{\hat{\mu}}$} from the $n^2$ template candidates as a mask to filter out noisy activations from $x$. \emph{I.e.} we compute {\small$\hat{\mu}\!=\!{\arg\!\max}_{[i,j]}x_{ij}$} and {\small$x^{\textrm{masked}}\!=\!\max\{x\circ T_{\hat{\mu}},0\}$}, where $\circ$ denotes the Hadamard (element-wise) product. {\small$\mu\!=\![i,j]$}, {\small$1\!\leq\!i,j\!\leq\!n$} denotes the unit (or location) in $x$ potentially corresponding to the part.

\begin{figure}[t]
\centering
\includegraphics[width=0.99\linewidth]{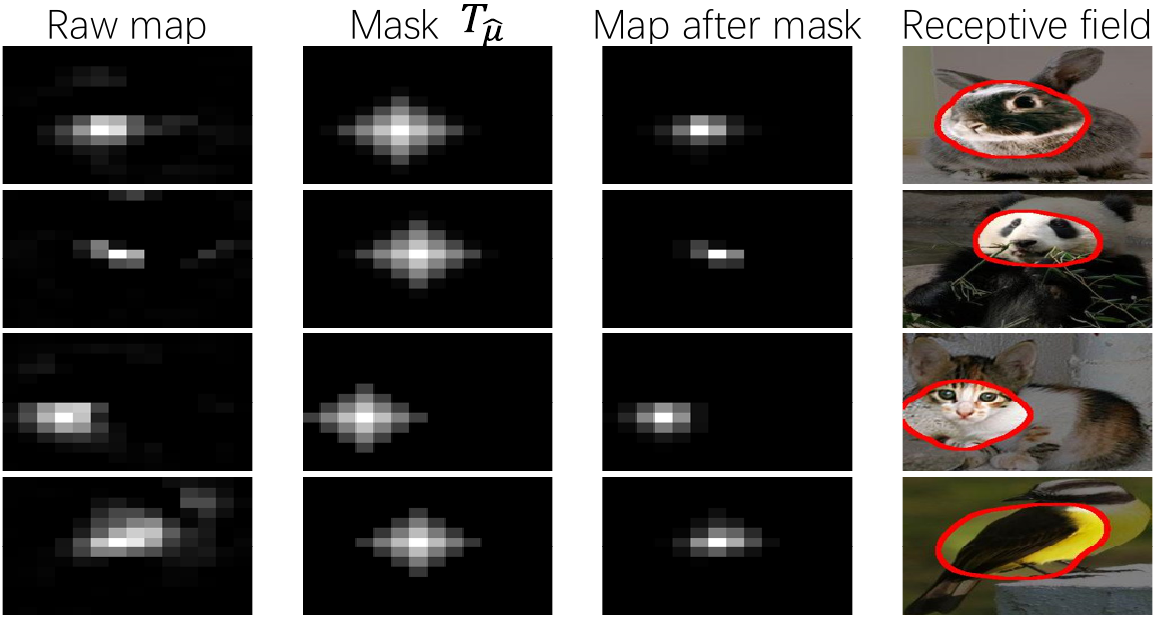}
\caption{Given an input image $I$, from the left to the right, we consequently show the feature map of a filter after the ReLU layer $x$, the assigned mask {\small$T_{\hat{\mu}}$}, the masked feature map {\small$x^{\textrm{masked}}$}, and the image-resolution RF of activations in {\small$x^{\textrm{masked}}$} computed by \cite{CNNSemanticDeep}.}
\label{fig:map}
\vspace{-8pt}
\end{figure}

The mask operation supports the gradient back-propagation for end-to-end learning. Note that the CNN may select different templates for different input images. Fig.~\ref{fig:map} visualizes the masks {\small$T_{\hat{\mu}}$} chosen for different images, as well as the original and masked feature maps.

\textbf{During the back-propagation process}, our loss pushes filter $f$ to represent a specific object part of the category $c$ and keep silent on images of other categories. Please see Section~\ref{sec:learning} for the determination of the category $c$ for filter $f$. Let {\small${\bf X}=\{x|x=f(I),I\in{\bf I}\}$} denote feature maps of $f$ after an ReLU operation, which are computed on different training images. Given an input image $I$, if {\small$I\in{\bf I}_{c}$}, we expect the feature map $x=f(I)$ to exclusively activated at the target part's location; otherwise, the feature map keeps inactivated. In other words, if {\small$I\in{\bf I}_{c}$}, the feature map $x$ is expected to the assigned template {\small$T_{\hat{\mu}}$}; if {\small$I\not\in{\bf I}_{c}$}, we design a negative template {\small$T^{-}$} and hope the feature map $x$ matches to {\small$T^{-}$}. Note that during the forward propagation, our method omits the negative template, and all feature maps, including those of other categories, select positive templates as masks.

Thus, each feature map is supposed to be well fit to one of all the {\small$n^2+1$} template candidates {\small${\bf T}=\{T^{-},T_{\mu_1},T_{\mu_2},\ldots,T_{\mu_{n^2}}\}$}. We formulate the loss for $f$ as the mutual information between {\small${\bf X}$} and {\small${\bf T}$}.
\begin{equation}
\begin{split}
{\bf Loss}_{f}=&-MI({\bf X};{\bf T})\quad\textrm{for filter }f\\
=&-\sum_{T}p(T)\sum_{x}p(x|T)\log\frac{p(x|T)}{p(x)}
\end{split}
\label{eqn:loss}
\vspace{-2pt}
\end{equation}
The prior probability of a template is given as {\small$p(T_{\mu})\!=\!\frac{\alpha}{n^2}, p(T^{-})\!=\!1-\alpha$}, where $\alpha$ is a constant prior likelihood. The fitness between a feature map $x$ and a template {\small$T$} is measured as the conditional likelihood {\small$p(x|T)$}.
\begin{equation}
\forall T\in{\bf T},\qquad p(x|T)=\frac{1}{Z_{T}}\exp\big[tr(x\cdot T)\big]
\vspace{-2pt}
\end{equation}
where {\small$Z_{T}=\sum_{x\in{\bf X}}\exp(tr(x\cdot T))$}. {\small$x\cdot T$} indicates the multiplication between $x$ and {\small$T$}; {\small$tr(\cdot)$} indicates the trace of a matrix, and {\small$tr(x\cdot T)=\sum_{ij}x_{ij}t_{ij}$}. {\small$p(x)=\sum_{T}p(T)p(x|T)$}.

\textbf{Part templates:} As shown in Fig.~\ref{fig:template}, a negative template is given as {\small$T^{-}\!=\!(t^{-}_{ij})$}, {\small$t^{-}_{ij}\!=\!-\tau\!<\!0$}, where $\tau$ is a positive constant. A positive template corresponding to $\mu$ is given as {\small$T_{\mu}\!=\!(t^{+}_{ij})$}, {\small$t^{+}_{ij}\!=\!\tau\cdot\max(1-\beta\frac{\Vert[i,j]-\mu\Vert_1}{n},-1)$}, where {\small$\Vert\cdot\Vert_1$} denotes the L-1 norm distance; $\beta$ is a constant parameter.

\subsection{Learning}
\label{sec:learning}

We train the interpretable CNN via an end-to-end manner. During the forward-propagation process, each filter in the CNN passes its information in a bottom-up manner, just like traditional CNNs. During the back-propagation process, each filter in an interpretable conv-layer receives gradients \emph{w.r.t.} its feature map $x$ from both the final task loss {\small${\bf L}(\hat{y}_{k},y^{*}_{k})$} and the local filter loss {\small${\bf Loss}_{f}$}, as follows:
\begin{equation}
\frac{\partial{\bf Loss}}{\partial x_{ij}}=\lambda\frac{\partial{\bf Loss}_{f}}{\partial x_{ij}}+\frac{1}{N}\sum_{i=k}^{N}\frac{\partial{\bf L}(\hat{y}_{k},y^{*}_{k})}{\partial x_{ij}}
\vspace{-2pt}
\end{equation}
where $\lambda$ is a weight.

We compute gradients of {\small${\bf Loss}_{f}$} \emph{w.r.t.} each element {\small$x_{ij}$} of feature map $x$ as follows\footnote[2]{Please see the proof in the Appendix.}.
\begin{small}
\vspace{-2pt}
\begin{eqnarray}
\!\!\frac{\partial{\bf Loss}_{f}}{\partial x_{ij}}\!=\!\frac{1}{Z_{T}}\!\sum_{T}p(T)t_{ij}e^{tr(x\cdot T)}\!\Big\{tr(x\cdot T)\!-\!\log\big[Z_{T}p(x)\big]\!\Big\}\!\!\!\!\!\!\!\!\!\nonumber\\
\approx\frac{p(\hat{T})\hat{t}_{ij}}{Z_{\hat{T}}}e^{tr(x\cdot\hat{T})}\Big\{tr(x\cdot\hat{T})-\log{Z_{\hat{T}}}-\log{p(x)}\Big\}
\label{eqn:grad}
\vspace{-2pt}
\end{eqnarray}
\end{small}
where {\small$\hat{T}$} is the target template for feature map $x$. If the given image $I$ belongs to the target category of filter $f$, then {\small$\hat{T}\!=\!T_{\hat{\mu}}$}, where {\small$\hat{\mu}\!=\!{\arg\!\max}_{[i,j]}x_{ij}$}. If image $I$ belongs to other categories, then {\small$\hat{T}\!=\!T^{-}$}. Considering {\small$\forall T\!\in\!{\bf T}\setminus\{\hat{T}\}$}, {\small$e^{tr(x\cdot\hat{T})}\!\gg\!e^{tr(x\cdot T)}$} after initial learning episodes, we make the above approximation to simplify the computation. Because {\small$Z_{T}$} is computed using numerous feature maps, we can roughly treat {\small$Z_{T}$} as a constant to compute gradients computation in the above equation. We gradually update the value of {\small$Z_{T}$} during the training process\footnote[3]{We can use a subset of feature maps to approximate the value of $Z_{T}$, and continue to update $Z_{T}$ when we receive more feature maps during the training process. Similarly, we can approximate $p(x)$ using a subset of feature maps. We compute $p(x)\!=\!\sum_{T}p(T)p(x|T)\!=\!\sum_{T}p(T)\frac{\exp[tr(x\cdot T)]}{Z_{T}}\approx\sum_{T}p(T)\textrm{mean}_{x}\frac{\exp[tr(x\cdot T)]}{Z_{T}}$.}. Similarly, we can also approximate {\small$p(x)$} without huge computation\footnotemark[3].

\textbf{Determining the target category for each filter:} We need to assign each filter $f$ with a target category $\hat{c}$ to approximate gradients in Eqn.~(\ref{eqn:grad}). We simply assign the filter $f$ with the category $\hat{c}$ whose images activate $f$ most, \emph{i.e.} {\small$\hat{c}={\arg\!\max}_{c}\textrm{mean}_{x=f(I):I\in{\bf I}_{c}}\sum_{ij}x_{ij}$}.

\section{Understanding of the loss}

In fact, the loss in Eqn.~(\ref{eqn:loss}) can be re-written as\footnotemark[2]
\begin{equation}
\begin{split}
{\bf Loss}_{f}=&-H({\bf T})+H({\bf T}'=\{T^{-},{\bf T}^{+}\}|{\bf X})\\
&+\sum_{x}p({\bf T}^{+},x)H({\bf T}^{+}|X=x)
\end{split}
\label{eqn:understand}
\vspace{-2pt}
\end{equation}
In the above equation, the first term {\small$H({\bf T})=-\sum_{T\in{\bf T}}p(T)\log p(T)$} is a constant, which denotes the prior entropy of part templates.

\noindent
\textbf{Low inter-category entropy:} The second term {\small$H({\bf T}'\!=\!\{T^{-}\!,\!{\bf T}^{+}\}|{\bf X})$} is computed as
\begin{equation}
\!\!\!H({\bf T}'\!=\!\{T^{-}\!,\!{\bf T}^{+}\}|{\bf X})=-\sum_{x}p(x)\!\!\!\!\!\!\!\!\!\sum_{T\in\{T^{-}\!,\!{\bf T}^{+}\}}\!\!\!\!\!\!\!\!\!p(T|x)\log p(T|x)\!
\vspace{-2pt}
\end{equation}
where {\small${\bf T}^{+}=\{T_{\mu_{1}},T_{\mu_{2}},\ldots,T_{\mu_{n^2}}\}\subset{\bf T}$}, {\small$p({\bf T}^{+}|x)=\sum_{\mu}p(T_{\mu}|x)$}. This term encourages a low conditional entropy of inter-category activations, \emph{i.e.} a well-learned filter $f$ needs to be exclusively activated by a certain category $c$ and keep silent on other categories. We can use a feature map $x$ of $f$ to identify whether the input image belongs to category $c$ or not, \emph{i.e.} $x$ fitting to either {\small$T_{\hat{\mu}}$} or {\small$T^{-}$}, without great uncertainty. Here, we define the set of all positive templates {\small${\bf T}^{+}$} as a single label to represent category $c$. We use the negative template {\small$T^{-}$} to denote other categories.

\noindent
\textbf{Low spatial entropy:} The third term in Eqn.~(\ref{eqn:understand}) is given as
\begin{equation}
H({\bf T}^{+}|X\!=\!x)=\sum_{\mu}\tilde{p}(T_{\mu}|x)\log\tilde{p}(T_{\mu}|x)
\vspace{-2pt}
\end{equation}
where {\small$\tilde{p}(T_{\mu}|x)=\frac{p(T_{\mu}|x)}{p({\bf T}^{+}|x)}$}. This term encourages a low conditional entropy of spatial distribution of $x$'s activations. \emph{I.e.} given an image {\small$I\in{\bf I}_{c}$}, a well-learned filter should only be activated by a single region $\hat{\mu}$ of the feature map $x$, instead of repetitively appearing at different locations.

\section{Experiments}

In experiments, to demonstrate the broad applicability, we applied our method to CNNs with four types of structures. We used object images in three different benchmark datasets to learn interpretable CNNs for single-category classification and multi-category classification. We visualized feature maps of filters in interpretable conv-layers to illustrate semantic meanings of these filters. We used two types of metrics, \emph{i.e.} the object-part interpretability and the location stability, to evaluate the clarity of the part semantics of a convolutional filter. Experiments showed that filters in our interpretable CNNs were much more semantically meaningful than those in ordinary CNNs.

\textbf{Three benchmark datasets:} Because we needed ground-truth annotations of object landmarks\footnote[4]{To avoid ambiguity, a landmark is referred to as the \textit{central position} of a semantic part (a part with an explicit name, \emph{e.g.} a head, a tail). In contrast, the part corresponding to a filter does not have an explicit name.} (parts) to evaluate the semantic clarity of each filter, we chose three benchmark datasets with landmark\footnotemark[4]/part annotations for training and testing, including the ILSVRC 2013 DET Animal-Part dataset~\cite{CNNAoG}, the CUB200-2011 dataset~\cite{CUB200}, and the Pascal VOC Part dataset~\cite{SemanticPart}. As discussed in \cite{SemanticPart,CNNAoG}, non-rigid parts of animal categories usually present great challenges for part localization. Thus, we followed \cite{SemanticPart,CNNAoG} to select the 37 animal categories in the three datasets for evaluation.

All the three datasets provide ground-truth bounding boxes of entire objects. For landmark annotations, the ILSVRC 2013 DET Animal-Part dataset~\cite{CNNAoG} contains ground-truth bounding boxes of heads and legs of 30 animal categories. The CUB200-2011 dataset~\cite{CUB200} contains a total of 11.8K bird images of 200 species, and the dataset provides center positions of 15 bird landmarks. The Pascal VOC Part dataset~\cite{SemanticPart} contain ground-truth part segmentations of 107 object landmarks in six animal categories.

\textbf{Four types of CNNs:} To demonstrate the broad applicability of our method, we modified four typical CNNs, \emph{i.e.} the AlexNet~\cite{CNNImageNet}, the VGG-M~\cite{VGG}, the VGG-S~\cite{VGG}, the VGG-16~\cite{VGG}, into interpretable CNNs. Considering that skip connections in residual networks~\cite{ResNet} usually make a single feature map encode patterns of different filters, in this study, we did not test the performance on residual networks to simplify the story. Given a certain CNN structure, we modified all filters in the top conv-layer of the original network into interpretable ones. Then, we inserted a new conv-layer with {\small$M$} filters above the original top conv-layer, where {\small$M$} is the channel number of the input of the new conv-layer. We also set filters in the new conv-layer as interpretable ones. Each filter was a {\small$3\times3\times M$} tensor with a bias term. We added zero padding to input feature maps to ensure that output feature maps were of the same size as the input.

\textbf{Implementation details:} We set parameters as {\small$\tau=\frac{0.5}{n^2}$}, {\small$\alpha=\frac{n^2}{1+n^2}$}, and {\small$\beta=4$}. We updated weights of filter losses \emph{w.r.t.} magnitudes of neural activations in an online manner, {\small$\lambda=5\times10^{-6}{\textrm{mean}}_{x\in{\bf X}}\max_{i,j}x_{ij}$}. We initialized parameters of fully-connected (FC) layers and the new conv-layer, and loaded parameters of other conv-layers from a traditional CNN that was pre-trained using 1.2M ImageNet images in \cite{CNNImageNet,VGG}. We then fine-tuned the interpretable CNN using training images in the dataset. To enable a fair comparison, traditional CNNs were also fine-tuned by initializing FC-layer parameters and loading conv-layer parameters.

\subsection{Experiments}

\textbf{Single-category classification:} We learned four types of interpretable CNNs based on the AlexNet, VGG-M, VGG-S, and VGG-16 structures to classify each category in the ILSVRC 2013 DET Animal-Part dataset~\cite{CNNAoG}, the CUB200-2011 dataset~\cite{CUB200}, and the Pascal VOC Part dataset~\cite{SemanticPart}. Besides, we also learned ordinary AlexNet, VGG-M, VGG-S, and VGG-16 networks using the same training data for comparison. We used the logistic log loss for single-category classification. Following experimental settings in \cite{CNNAoG,DeepQA,explanatoryGraph}, we cropped objects of the target category based on their bounding boxes as positive samples with ground-truth labels {\small$y^{*}\!=\!+1$}. We regarded images of other categories as negative samples with ground-truth labels {\small$y^{*}\!=\!-1$}.

\textbf{Multi-category classification:} We used the six animal categories in the Pascal VOC Part dataset~\cite{SemanticPart} and the thirty categories in the ILSVRC 2013 DET Animal-Part dataset~\cite{CNNAoG} respectively, to learn CNNs for multi-category classification. We learned interpretable CNNs based on the VGG-M, VGG-S, and VGG-16 structures. We tried two types of losses, \emph{i.e.} the softmax log loss and the logistic log loss\footnote[5]{We considered the output $y_{c}$ for each category $c$ independent to outputs for other categories, thereby a CNN making multiple independent single-class classifications for each image. Table~\ref{tab:classification} reported the average accuracy of the multiple classification outputs of an image.} for multi-class classification.

\subsection{Quantitative evaluation of part interpretability}

As discussed in \cite{Interpretability}, filters in low conv-layers usually represent simple patterns or object details (\emph{e.g.} edges, simple textures, and colors), whereas filters in high conv-layers are more likely to represent complex, large-scale parts. Therefore, in experiments, we evaluated the clarity of part semantics for the top conv-layer of a CNN. We used the following two metrics for evaluation.


\subsubsection{Evaluation metric: part interpretability}
\label{sec:interpretability}

\begin{figure*}[t]
\centering
\includegraphics[width=0.99\linewidth]{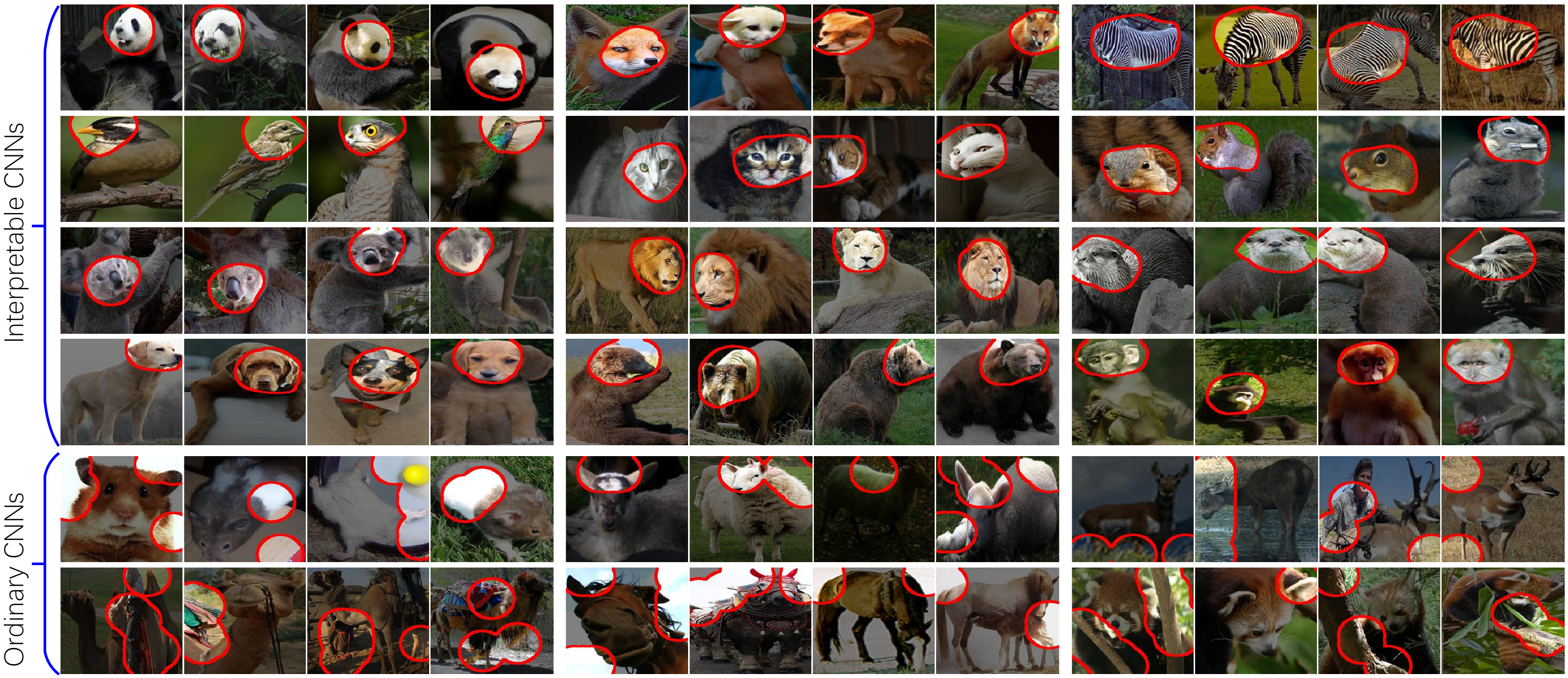}
\caption{Visualization of filters in top conv-layers. We used \cite{CNNSemanticDeep} to estimate the image-resolution receptive field of activations in a feature map to visualize a filter's semantics. The top four rows visualize filters in interpretable CNNs, and the bottom two rows correspond to filters in ordinary CNNs. We found that interpretable CNNs usually encoded head patterns of animals in its top conv-layer for classification.}
\label{fig:visual}
\vspace{-10pt}
\end{figure*}

We followed the metric proposed by Bau~\emph{et al.}~\cite{Interpretability} to measure the object-part interpretability of filters. We briefly introduce this evaluation metric as follows. For each filter $f$, we computed its feature maps {\small${\bf X}$} after ReLu/mask operations on different input images. Then, the distribution of activation scores in all positions of all feature maps was computed. \cite{Interpretability} set an activation threshold {\small$T_{f}$} such that {\small$p(x_{ij}>T_{f})=0.005$}, so as to select top activations from all spatial locations {\small$[i,j]$} of all feature maps {\small$x\in{\bf X}$} as valid map regions corresponding to $f$'s semantics. Then, \cite{Interpretability} scaled up low-resolution valid map regions to the image resolution, thereby obtaining the receptive field (RF)\footnote[6]{Note that \cite{CNNSemanticDeep} accurately computes the RF when the filter represents an object part, and we used RFs computed by \cite{CNNSemanticDeep} for filter visualization in Fig.~\ref{fig:visual}. However, when a filter in an ordinary CNN does not have consistent contours, it is difficult for \cite{CNNSemanticDeep} to align different images to compute an average RF. Thus, for ordinary CNNs, we simply used a round RF for each valid activation. We overlapped all activated RFs in a feature map to compute the final RF as mentioned in \cite{Interpretability}. For a fair comparison, in Section~\label{sec:interpretability}, we uniformly applied these RFs to both interpretable CNNs and ordinary CNNs.} of valid activations on each image. The RF on image $I$, denoted by {\small$S_{f}^{I}$}, described the part region of $f$.

\begin{table}
\resizebox{1.0\linewidth}{!}{\begin{tabular}{c|ccccccc}
\hline
\!\!\!&\!\! bird \!\!\!&\!\! cat \!\!\!&\!\! cow \!\!\!&\!\! dog \!\!\!&\!\! {\small horse} \!\!\!&\!\! {\small sheep} \!\!\!&\!\! \textcolor{blue}{\bf Avg.}\\
\hline
\!\!\!AlexNet &\!\!\!
0.332\!\!\!&\!\!
0.363\!\!\!&\!\!
0.340\!\!\!&\!\!
0.374\!\!\!&\!\!
0.308\!\!\!&\!\!
0.373\!\!\!&\!\!
\textcolor{blue}{0.348}\!\!\!
\\
\!\!\!{\footnotesize AlexNet, interpretable} &\!\!\!
{\bf0.770}\!\!\!&\!\!
{\bf0.565}\!\!\!&\!\!
{\bf0.618}\!\!\!&\!\!
{\bf0.571}\!\!\!&\!\!
{\bf0.729}\!\!\!&\!\!
{\bf0.669}\!\!\!&\!\!
\textcolor{blue}{{\bf0.654}}\!\!\!
\\
\hline
\!\!\!VGG-16 &\!\!\!
0.519\!\!\!&\!\!
0.458\!\!\!&\!\!
0.479\!\!\!&\!\!
0.534\!\!\!&\!\!
0.440\!\!\!&\!\!
0.542\!\!\!&\!\!
\textcolor{blue}{0.495}\!\!\!
\\
\!\!\!{\footnotesize VGG-16, interpretable} &\!\!\!
{\bf0.818}\!\!\!&\!\!
{\bf0.653}\!\!\!&\!\!
{\bf0.683}\!\!\!&\!\!
{\bf0.900}\!\!\!&\!\!
{\bf0.795}\!\!\!&\!\!
{\bf0.772}\!\!\!&\!\!
\textcolor{blue}{{\bf0.770}}\!\!\!
\\
\hline
\!\!\!VGG-M &\!\!\!
0.357\!\!\!&\!\!
0.365\!\!\!&\!\!
0.347\!\!\!&\!\!
0.368\!\!\!&\!\!
0.331\!\!\!&\!\!
0.373\!\!\!&\!\!
\textcolor{blue}{0.357}\!\!\!
\\
\!\!\!{\footnotesize VGG-M, interpretable} &\!\!\!
{\bf0.821}\!\!\!&\!\!
{\bf0.632}\!\!\!&\!\!
{\bf0.634}\!\!\!&\!\!
{\bf0.669}\!\!\!&\!\!
{\bf0.736}\!\!\!&\!\!
{\bf0.756}\!\!\!&\!\!
\textcolor{blue}{{\bf0.708}}\!\!\!
\\
\hline
\!\!\!VGG-S &\!\!\!
0.251\!\!\!&\!\!
0.269\!\!\!&\!\!
0.235\!\!\!&\!\!
0.275\!\!\!&\!\!
0.223\!\!\!&\!\!
{\bf0.287}\!\!\!&\!\!
\textcolor{blue}{0.257}\!\!\!
\\
\!\!\!{\footnotesize VGG-S, interpretable} &\!\!\!
{\bf0.526}\!\!\!&\!\!
{\bf0.366}\!\!\!&\!\!
{\bf0.291}\!\!\!&\!\!
{\bf0.432}\!\!\!&\!\!
{\bf0.478}\!\!\!&\!\!
0.251\!\!\!&\!\!
\textcolor{blue}{{\bf0.390}}\!\!\!
\\
\hline
\end{tabular}}
\vspace{1pt}
\caption{Average part interpretability of filters in CNNs for single-category classification using the Pascal VOC Part dataset~\cite{SemanticPart}.}
\label{tab:interpretability}
\vspace{-8pt}
\end{table}

The compatibility between each filter $f$ and the $k$-th part on image $I$ was reported as an intersection-over-union score {\small$IoU_{f,k}^{I}\!=\!\frac{\Vert S_{f}^{I}\cap S_{k}^{I}\Vert}{\Vert S_{f}^{I}\cup S_{k}^{I}\Vert}$}, where {\small$S_{k}^{I}$} denotes the ground-truth mask of the $k$-th part on image $I$. Given an image $I$, we associated filter $f$ with the $k$-th part if {\small$IoU_{f,k}^{I}>0.2$}. Note that the criterion of {\small$IoU_{f,k}^{I}>0.2$} for part association is much stricter than {\small$IoU_{f,k}^{I}>0.04$} that was used in \cite{Interpretability}. It is because compared to other CNN semantics discussed in \cite{Interpretability} (such as colors and textures), object-part semantics requires a stricter criterion. We computed the probability of the $k$-th part being associating with the filter $f$ as {\small$P_{f,k}={\textrm{mean}}_{I:\textrm{with k-th part}}{\bf1}(IoU_{f,k}^{I}>0.2)$}. Note that one filter might be associated with multiple object parts in an image. Among all parts, we reported the highest probability of part association as the interpretability of filter $f$, \emph{i.e.} {\small$P_{f}=\max_{k}P_{f,k}$}.

\textbf{For single-category classification,} we used testing images of the target category for evaluation. In the Pascal VOC Part dataset~\cite{SemanticPart}, we used four parts for the \textit{bird} category. We merged ground-truth regions of the head, beak, and l/r-eyes as the head part, merged regions of the torso, neck, and l/r-wings as the torso part, merged regions of l/r-legs/feet as the leg part, and used tail regions as the fourth part. We used five parts for the \textit{cat} category. We merged regions of the head, l/r-eyes, l/r-ears, and nose as the head part, merged regions of the torso and neck as the torso part, merged regions of frontal l/r-legs/paws as the frontal legs, merged regions of back l/r-legs/paws as the back legs, and used the tail as the fifth part. We used four parts for the \textit{cow} category, which were defined in a similar way to the cat category. We added l/r-horns to the head part and omitted the tail part. We applied five parts of the \textit{dog} category in the same way as the cat category. We applied four parts of both the \textit{horse} and \textit{sheep} categories in the same way as the cow category. We computed the average part interpretability {\small$P_{f}$} over all filters for evaluation.

\textbf{For multi-category classification,} we first assigned each filter $f$ with a target category $\hat{c}$, \emph{i.e.} the category that activated the filter most {\small$\hat{c}\!=\!{\arg\!\max}_{c}{\textrm{mean}}_{x:I\in{\bf I}_{c}}\sum_{i,j}x_{ij}$}. Then, we computed the object-part interpretability using images of category $\hat{c}$, as introduced above.

\begin{table}[t]
\centering
\resizebox{1.0\linewidth}{!}{\begin{tabular}{ccc}
\hline
\qquad Network\quad & \quad Logistic log loss\footnotemark[5]\quad & \quad Softmax log loss\qquad\\
\hline
VGG-16 &0.710 &0.723\\
{\small VGG-16, interpretable} &{\bf0.938} & {\bf0.897}\\
\hline
VGG-M &0.478 &0.502\\
{\small VGG-M, interpretable} &{\bf0.770} &{\bf0.734}\\
\hline
VGG-S &0.479 &0.435\\
{\small VGG-S, interpretable} &{\bf0.572} &{\bf0.601}\\
\hline
\end{tabular}}
\vspace{1pt}
\caption{Average part interpretability of filters in CNNs that are trained for multi-category classification. Filters in our interpretable CNNs exhibited significantly better part interpretability than other CNNs in all comparisons.}
\label{tab:multi-interpretability}
\vspace{-8pt}
\end{table}

\subsubsection{Evaluation metric: location stability}

\begin{table*}[t]
\centering
\resizebox{1.0\linewidth}{!}{\begin{tabular}{p{2.5cm}|cccccccccccccccc}
\hline
\!\!\!&\!\! gold. \!\!\!&\!\! bird \!\!\!&\!\! frog \!\!\!&\!\! turt. \!\!\!&\!\! liza. \!\!\!&\!\! koala \!\!\!&\!\! lobs. \!\!\!&\!\! dog \!\!\!&\!\! fox \!\!\!&\!\! cat \!\!\!&\!\! lion \!\!\!&\!\! tiger \!\!\!&\!\! bear \!\!\!&\!\! rabb. \!\!\!&\!\! hams. \!\!\!&\!\! squi.\\
\!\!\! AlexNet\!\!\!&\!\!
0.161\!\!\!&\!\!
0.167\!\!\!&\!\!
0.152\!\!\!&\!\!
0.153\!\!\!&\!\!
0.175\!\!\!&\!\!
0.128\!\!\!&\!\!
0.123\!\!\!&\!\!
0.144\!\!\!&\!\!
0.143\!\!\!&\!\!
0.148\!\!\!&\!\!
0.137\!\!\!&\!\!
0.142\!\!\!&\!\!
0.144\!\!\!&\!\!
0.148\!\!\!&\!\!
0.128\!\!\!&\!\!
0.149
\\
\!\!\! {\footnotesize AlexNet, interpretable}\!\!\!&\!\!
{\bf0.084}\!\!\!&\!\!
{\bf0.095}\!\!\!&\!\!
{\bf0.090}\!\!\!&\!\!
{\bf0.107}\!\!\!&\!\!
{\bf0.097}\!\!\!&\!\!
{\bf0.079}\!\!\!&\!\!
{\bf0.077}\!\!\!&\!\!
{\bf0.093}\!\!\!&\!\!
{\bf0.087}\!\!\!&\!\!
{\bf0.095}\!\!\!&\!\!
{\bf0.084}\!\!\!&\!\!
{\bf0.090}\!\!\!&\!\!
{\bf0.095}\!\!\!&\!\!
{\bf0.095}\!\!\!&\!\!
{\bf0.077}\!\!\!&\!\!
{\bf0.095}
\\
\cline{1-1}
\!\!\! VGG-16\!\!\!&\!\!
0.153\!\!\!&\!\!
0.156\!\!\!&\!\!
0.144\!\!\!&\!\!
0.150\!\!\!&\!\!
0.170\!\!\!&\!\!
0.127\!\!\!&\!\!
0.126\!\!\!&\!\!
0.143\!\!\!&\!\!
0.137\!\!\!&\!\!
0.148\!\!\!&\!\!
0.139\!\!\!&\!\!
0.144\!\!\!&\!\!
0.143\!\!\!&\!\!
0.146\!\!\!&\!\!
0.125\!\!\!&\!\!
0.150
\\
\!\!\! {\footnotesize VGG-16, interpretable}\!\!\!&\!\!
{\bf0.076}\!\!\!&\!\!
{\bf0.099}\!\!\!&\!\!
{\bf0.086}\!\!\!&\!\!
{\bf0.115}\!\!\!&\!\!
{\bf0.113}\!\!\!&\!\!
{\bf0.070}\!\!\!&\!\!
{\bf0.084}\!\!\!&\!\!
{\bf0.077}\!\!\!&\!\!
{\bf0.069}\!\!\!&\!\!
{\bf0.086}\!\!\!&\!\!
{\bf0.067}\!\!\!&\!\!
{\bf0.097}\!\!\!&\!\!
{\bf0.081}\!\!\!&\!\!
{\bf0.079}\!\!\!&\!\!
{\bf0.066}\!\!\!&\!\!
{\bf0.065}
\\
\cline{1-1}
\!\!\! VGG-M\!\!\!&\!\!
0.161\!\!\!&\!\!
0.166\!\!\!&\!\!
0.151\!\!\!&\!\!
0.153\!\!\!&\!\!
0.176\!\!\!&\!\!
0.128\!\!\!&\!\!
0.125\!\!\!&\!\!
0.145\!\!\!&\!\!
0.145\!\!\!&\!\!
0.150\!\!\!&\!\!
0.140\!\!\!&\!\!
0.145\!\!\!&\!\!
0.144\!\!\!&\!\!
0.150\!\!\!&\!\!
0.128\!\!\!&\!\!
0.150
\\
\!\!\! {\footnotesize VGG-M, interpretable}\!\!\!&\!\!
{\bf0.088}\!\!\!&\!\!
{\bf0.088}\!\!\!&\!\!
{\bf0.089}\!\!\!&\!\!
{\bf0.108}\!\!\!&\!\!
{\bf0.099}\!\!\!&\!\!
{\bf0.080}\!\!\!&\!\!
{\bf0.074}\!\!\!&\!\!
{\bf0.090}\!\!\!&\!\!
{\bf0.082}\!\!\!&\!\!
{\bf0.103}\!\!\!&\!\!
{\bf0.079}\!\!\!&\!\!
{\bf0.089}\!\!\!&\!\!
{\bf0.101}\!\!\!&\!\!
{\bf0.097}\!\!\!&\!\!
{\bf0.082}\!\!\!&\!\!
{\bf0.095}
\\
\cline{1-1}
\!\!\! VGG-S\!\!\!&\!\!
0.158\!\!\!&\!\!
0.166\!\!\!&\!\!
0.149\!\!\!&\!\!
0.151\!\!\!&\!\!
0.173\!\!\!&\!\!
0.127\!\!\!&\!\!
0.124\!\!\!&\!\!
0.143\!\!\!&\!\!
0.142\!\!\!&\!\!
0.148\!\!\!&\!\!
0.138\!\!\!&\!\!
0.142\!\!\!&\!\!
0.143\!\!\!&\!\!
0.148\!\!\!&\!\!
0.128\!\!\!&\!\!
0.146
\\
\!\!\! {\footnotesize VGG-S, interpretable}\!\!\!&\!\!
{\bf0.087}\!\!\!&\!\!
{\bf0.101}\!\!\!&\!\!
{\bf0.093}\!\!\!&\!\!
{\bf0.107}\!\!\!&\!\!
{\bf0.096}\!\!\!&\!\!
{\bf0.084}\!\!\!&\!\!
{\bf0.078}\!\!\!&\!\!
{\bf0.091}\!\!\!&\!\!
{\bf0.082}\!\!\!&\!\!
{\bf0.101}\!\!\!&\!\!
{\bf0.082}\!\!\!&\!\!
{\bf0.089}\!\!\!&\!\!
{\bf0.097}\!\!\!&\!\!
{\bf0.091}\!\!\!&\!\!
{\bf0.076}\!\!\!&\!\!
{\bf0.098}
\\
\hline
\!\!\!&\!\! horse \!\!\!&\!\! zebra \!\!\!&\!\! swine \!\!\!&\!\! hippo \!\!\!&\!\! catt. \!\!\!&\!\! sheep \!\!\!&\!\! ante. \!\!\!&\!\! camel \!\!\!&\!\! otter \!\!\!&\!\! arma. \!\!\!&\!\! monk. \!\!\!&\!\! elep. \!\!\!&\!\! red pa. \!\!\!&\!\! gia.pa. \!\!\!&\!\! \!\!\!&\!\! \textcolor{blue}{\bf\large Avg.}\\
\!\!\! AlexNet\!\!\!&\!\!
0.152\!\!\!&\!\!
0.154\!\!\!&\!\!
0.141\!\!\!&\!\!
0.141\!\!\!&\!\!
0.144\!\!\!&\!\!
0.155\!\!\!&\!\!
0.147\!\!\!&\!\!
0.153\!\!\!&\!\!
0.159\!\!\!&\!\!
0.160\!\!\!&\!\!
0.139\!\!\!&\!\!
0.125\!\!\!&\!\!
0.140\!\!\!&\!\!
0.125\!\!\!&\!\!
\!\!\!&\!\!
\textcolor{blue}{0.146}
\\
\!\!\! {\footnotesize AlexNet, interpretable}\!\!\!&\!\!
{\bf0.098}\!\!\!&\!\!
{\bf0.084}\!\!\!&\!\!
{\bf0.091}\!\!\!&\!\!
{\bf0.089}\!\!\!&\!\!
{\bf0.097}\!\!\!&\!\!
{\bf0.101}\!\!\!&\!\!
{\bf0.085}\!\!\!&\!\!
{\bf0.102}\!\!\!&\!\!
{\bf0.104}\!\!\!&\!\!
{\bf0.095}\!\!\!&\!\!
{\bf0.090}\!\!\!&\!\!
{\bf0.085}\!\!\!&\!\!
{\bf0.084}\!\!\!&\!\!
{\bf0.073}\!\!\!&\!\!
\!\!\!&\!\!
\textcolor{blue}{\bf0.091}
\\
\cline{1-1}
\!\!\! VGG-16\!\!\!&\!\!
0.150\!\!\!&\!\!
0.153\!\!\!&\!\!
0.141\!\!\!&\!\!
0.140\!\!\!&\!\!
0.140\!\!\!&\!\!
0.150\!\!\!&\!\!
0.144\!\!\!&\!\!
0.149\!\!\!&\!\!
0.154\!\!\!&\!\!
0.163\!\!\!&\!\!
0.136\!\!\!&\!\!
0.129\!\!\!&\!\!
0.143\!\!\!&\!\!
0.125\!\!\!&\!\!
\!\!\!&\!\!
\textcolor{blue}{0.144}
\\
\!\!\! {\footnotesize VGG-16, interpretable}\!\!\!&\!\!
{\bf0.106}\!\!\!&\!\!
{\bf0.077}\!\!\!&\!\!
{\bf0.094}\!\!\!&\!\!
{\bf0.083}\!\!\!&\!\!
{\bf0.102}\!\!\!&\!\!
{\bf0.097}\!\!\!&\!\!
{\bf0.091}\!\!\!&\!\!
{\bf0.105}\!\!\!&\!\!
{\bf0.093}\!\!\!&\!\!
{\bf0.100}\!\!\!&\!\!
{\bf0.074}\!\!\!&\!\!
{\bf0.084}\!\!\!&\!\!
{\bf0.067}\!\!\!&\!\!
{\bf0.063}\!\!\!&\!\!
\!\!\!&\!\!
\textcolor{blue}{\bf0.085}
\\
\cline{1-1}
\!\!\! VGG-M\!\!\!&\!\!
0.151\!\!\!&\!\!
0.158\!\!\!&\!\!
0.140\!\!\!&\!\!
0.140\!\!\!&\!\!
0.143\!\!\!&\!\!
0.155\!\!\!&\!\!
0.146\!\!\!&\!\!
0.154\!\!\!&\!\!
0.160\!\!\!&\!\!
0.161\!\!\!&\!\!
0.140\!\!\!&\!\!
0.126\!\!\!&\!\!
0.142\!\!\!&\!\!
0.127\!\!\!&\!\!
\!\!\!&\!\!
\textcolor{blue}{0.147}
\\
\!\!\! {\footnotesize VGG-M, interpretable}\!\!\!&\!\!
{\bf0.095}\!\!\!&\!\!
{\bf0.080}\!\!\!&\!\!
{\bf0.095}\!\!\!&\!\!
{\bf0.084}\!\!\!&\!\!
{\bf0.092}\!\!\!&\!\!
{\bf0.094}\!\!\!&\!\!
{\bf0.077}\!\!\!&\!\!
{\bf0.104}\!\!\!&\!\!
{\bf0.102}\!\!\!&\!\!
{\bf0.093}\!\!\!&\!\!
{\bf0.086}\!\!\!&\!\!
{\bf0.087}\!\!\!&\!\!
{\bf0.089}\!\!\!&\!\!
{\bf0.068}\!\!\!&\!\!
\!\!\!&\!\!
\textcolor{blue}{\bf0.090}
\\
\cline{1-1}
\!\!\! VGG-S\!\!\!&\!\!
0.149\!\!\!&\!\!
0.155\!\!\!&\!\!
0.139\!\!\!&\!\!
0.140\!\!\!&\!\!
0.141\!\!\!&\!\!
0.155\!\!\!&\!\!
0.143\!\!\!&\!\!
0.154\!\!\!&\!\!
0.158\!\!\!&\!\!
0.157\!\!\!&\!\!
0.140\!\!\!&\!\!
0.125\!\!\!&\!\!
0.139\!\!\!&\!\!
0.125\!\!\!&\!\!
\!\!\!&\!\!
\textcolor{blue}{0.145}
\\
\!\!\! {\footnotesize VGG-S, interpretable}\!\!\!&\!\!
{\bf0.096}\!\!\!&\!\!
{\bf0.080}\!\!\!&\!\!
{\bf0.092}\!\!\!&\!\!
{\bf0.088}\!\!\!&\!\!
{\bf0.094}\!\!\!&\!\!
{\bf0.101}\!\!\!&\!\!
{\bf0.077}\!\!\!&\!\!
{\bf0.102}\!\!\!&\!\!
{\bf0.105}\!\!\!&\!\!
{\bf0.094}\!\!\!&\!\!
{\bf0.090}\!\!\!&\!\!
{\bf0.086}\!\!\!&\!\!
{\bf0.078}\!\!\!&\!\!
{\bf0.072}\!\!\!&\!\!
\!\!\!&\!\!
\textcolor{blue}{\bf0.090}
\\
\hline
\end{tabular}}
\vspace{1pt}
\caption{Location instability of filters (${\bf E}_{f,k}[D_{f,k}]$) in CNNs that are trained for single-category classification using the ILSVRC 2013 DET Animal-Part dataset~\cite{CNNAoG}. Filters in our interpretable CNNs exhibited significantly lower localization instability than ordinary CNNs in all comparisons. \textcolor{blue}{Please see supplementary materials for performance of other structural modifications of CNNs.}}
\label{tab:imgnet-stability}
\end{table*}

The second metric measures the stability of part locations, which was proposed in \cite{explanatoryGraph}. Given a feature map $x$ of filter $f$, we regarded the unit $\hat{\mu}$ with the highest activation as the location inference of $f$. We assumed that if $f$ consistently represented the same object part through different objects, then distances between the inferred part location $\hat{\mu}$ and some object landmarks\footnotemark[4] should not change a lot among different objects. For example, if $f$ represented the shoulder, then the distance between the shoulder and the head should keep stable through different objects.

Therefore, \cite{explanatoryGraph} computed the deviation of the distance between the inferred position $\hat{\mu}$ and a specific ground-truth landmark among different images, and used the average deviation \emph{w.r.t.} various landmark to evaluate the location stability of $f$. A smaller deviation indicates a higher location stability. Let {\small$d_{I}(p_{k},\hat{\mu})=\frac{\Vert {\bf p}_{k}-{\bf p}(\hat{\mu})\Vert}{\sqrt{w^2+h^2}}$} denote the normalized distance between the inferred part and the $k$-th landmark {\small${\bf p}_{k}$} on image $I$, where {\small${\bf p}(\hat{\mu})$} denotes the center of the unit $\hat{\mu}$'s RF when we backward propagated the RF to the image plane. {\small$\sqrt{w^2+h^2}$} denotes the diagonal length of the input image. We computed {\small$D_{f,k}=\sqrt{{\textrm{var}}_{I}[d_{I}(p_{k},\hat{\mu})]}$} as the \textit{relative location deviation} of filter $f$ \emph{w.r.t.} the $k$-th landmark, where {\small${\textrm{var}}_{I}[d_{I}(p_{k},\hat{\mu})]$} is referred to as the variation of the distance {\small$d_{I}(p_{k},\hat{\mu})$}. Because each landmark could not appear in all testing images, for each filter $f$, we only used inference results with the top-100 highest activation scores {\small$x_{\hat{\mu}}$} on images containing the $k$-th landmark to compute {\small$D_{f,k}$}. Thus, we used the average of relative location deviations of all the filters in a conv-layer \emph{w.r.t.} all landmarks, \emph{i.e.} {\small${\textrm{mean}}_{f}{\textrm{mean}}_{k=1}^{K}D_{f,k}$}, to measure the location instability of $f$, where {\small$K$} denotes the number of landmarks.

More specifically, object landmarks for each category were selected as follows. For the ILSVRC 2013 DET Animal-Part dataset~\cite{CNNAoG}, we used the \textit{head} and \textit{frontal legs} of each category as landmarks for evaluation. For the Pascal VOC Part dataset~\cite{SemanticPart}, we selected the \textit{head}, \textit{neck}, and \textit{torso} of each category as the landmarks. For the CUB200-2011 dataset~\cite{CUB200}, we used ground-truth positions of the \textit{head}, \textit{back}, \textit{tail} of birds as landmarks. It was because these landmarks appeared on testing images most frequently.

\begin{table}[t]
\centering
\resizebox{1.0\linewidth}{!}{\begin{tabular}{c|ccccccc}
\hline
\!\!\!&\!\! bird \!\!\!&\!\! cat \!\!\!&\!\! cow \!\!\!&\!\! dog \!\!\!&\!\! {\small horse} \!\!\!&\!\! {\small sheep} \!\!\!&\!\! \textcolor{blue}{\bf Avg.}\\
\hline
\!\!\!AlexNet \!\!\!&\!\!
0.153\!\!\!&\!\!
0.131\!\!\!&\!\!
0.141\!\!\!&\!\!
0.128\!\!\!&\!\!
0.145\!\!\!&\!\!
0.140\!\!\!&\!\!
\textcolor{blue}{0.140}\!\!\!
\\
\!\!\!{\footnotesize AlexNet, interpretable} \!\!\!&\!\!
{\bf0.090}\!\!\!&\!\!
{\bf0.089}\!\!\!&\!\!
{\bf0.090}\!\!\!&\!\!
{\bf0.088}\!\!\!&\!\!
{\bf0.087}\!\!\!&\!\!
{\bf0.088}\!\!\!&\!\!
\textcolor{blue}{\bf0.088}\!\!\!
\\
\hline
\!\!\!VGG-16 \!\!\!&\!\!
0.145\!\!\!&\!\!
0.133\!\!\!&\!\!
0.146\!\!\!&\!\!
0.127\!\!\!&\!\!
0.143\!\!\!&\!\!
0.143\!\!\!&\!\!
\textcolor{blue}{0.139}\!\!\!
\\
\!\!\!{\footnotesize VGG-16, interpretable} \!\!\!&\!\!
{\bf0.101}\!\!\!&\!\!
{\bf0.098}\!\!\!&\!\!
{\bf0.105}\!\!\!&\!\!
{\bf0.074}\!\!\!&\!\!
{\bf0.097}\!\!\!&\!\!
{\bf0.100}\!\!\!&\!\!
\textcolor{blue}{\bf0.096}\!\!\!
\\
\hline
\!\!\!VGG-M \!\!\!&\!\!
0.152\!\!\!&\!\!
0.132\!\!\!&\!\!
0.143\!\!\!&\!\!
0.130\!\!\!&\!\!
0.145\!\!\!&\!\!
0.141\!\!\!&\!\!
\textcolor{blue}{0.141}\!\!\!
\\
\!\!\!{\footnotesize VGG-M, interpretable} \!\!\!&\!\!
{\bf0.086}\!\!\!&\!\!
{\bf0.094}\!\!\!&\!\!
{\bf0.090}\!\!\!&\!\!
{\bf0.087}\!\!\!&\!\!
{\bf0.084}\!\!\!&\!\!
{\bf0.084}\!\!\!&\!\!
\textcolor{blue}{\bf0.088}\!\!\!
\\
\hline
\!\!\!VGG-S \!\!\!&\!\!
0.152\!\!\!&\!\!
0.131\!\!\!&\!\!
0.141\!\!\!&\!\!
0.128\!\!\!&\!\!
0.144\!\!\!&\!\!
0.141\!\!\!&\!\!
\textcolor{blue}{0.139}\!\!\!
\\
\!\!\!{\footnotesize VGG-S, interpretable} \!\!\!&\!\!
{\bf0.089}\!\!\!&\!\!
{\bf0.092}\!\!\!&\!\!
{\bf0.092}\!\!\!&\!\!
{\bf0.087}\!\!\!&\!\!
{\bf0.086}\!\!\!&\!\!
{\bf0.088}\!\!\!&\!\!
\textcolor{blue}{\bf0.089}\!\!\!
\\
\hline
\end{tabular}}
\vspace{1pt}
\caption{Location instability of filters (${\bf E}_{f,k}[D_{f,k}]$) in CNNs that are trained for single-category classification using the Pascal VOC Part dataset~\cite{SemanticPart}. Filters in our interpretable CNNs exhibited significantly lower localization instability than ordinary CNNs in all comparisons. \textcolor{blue}{Please see supplementary materials for performance of other structural modifications of CNNs.}}
\label{tab:voc-stability}
\end{table}

\begin{table}[t]
\centering
\resizebox{1.0\linewidth}{!}{\begin{tabular}{cc}
\hline
\qquad\qquad Network\qquad\qquad\qquad & \qquad\qquad Avg. location instability\qquad\\
\hline
AlexNet &0.150\\
AlexNet, interpretable &{\bf0.070}\\
\hline
VGG-16 &0.137\\
VGG-16, interpretable &{\bf0.076}\\
\hline
VGG-M &0.148\\
VGG-M, interpretable &{\bf0.065}\\
\hline
VGG-S &0.148\\
VGG-S, interpretable &{\bf0.073}\\
\hline
\end{tabular}}
\vspace{1pt}
\caption{Location instability of filters (${\bf E}_{f,k}[D_{f,k}]$) in CNNs for single-category classification based on the CUB200-2011 dataset~\cite{CUB200}. \textcolor{blue}{Please see supplementary materials for performance of other structural modifications on ordinary CNNs.}}
\label{tab:cub200-stability}
\end{table}

\begin{table}[t]
\centering
\resizebox{1.0\linewidth}{!}{\begin{tabular}{c|c|cc}
\hline
Dataset & ILSVRC Part~\cite{CNNAoG} & \multicolumn{2}{|c}{Pascal VOC Part~\cite{SemanticPart}}\\
Network & {\footnotesize Logistic log loss\footnotemark[5]} & {\footnotesize Logistic log loss\footnotemark[5]} & {\footnotesize Softmax log loss}\\
\hline
VGG-16 & -- &0.128 &0.142\\
{\small interpretable} & -- &{\bf0.073} &{\bf0.075}\\
\hline
VGG-M & 0.167 &0.135 &0.137\\
{\small interpretable} &{\bf0.096} &{\bf0.083} &{\bf 0.087}\\
\hline
VGG-S & 0.131 &0.138 &0.138\\
{\small interpretable} &{\bf0.083} &{\bf0.078} &{\bf0.082}\\
\hline
\end{tabular}}
\vspace{1pt}
\caption{Location instability of filters (${\bf E}_{f,k}[D_{f,k}]$) in CNNs that are trained for multi-category classification. Filters in our interpretable CNNs exhibited significantly lower localization instability than ordinary CNNs in all comparisons.}
\label{tab:multi-stability}
\end{table}

For multi-category classification, we needed to determine two terms for each filter $f$, \emph{i.e.} 1) the category that $f$ mainly represented and 2) the relative location deviation {\small$D_{f,k}$} \emph{w.r.t.} landmarks in $f$'s target category. Because filters in ordinary CNNs did not exclusively represent a single category, we simply assigned filter $f$ with the category whose landmarks can achieve the lowest location deviation to simplify the computation. \emph{I.e.} we used the average location deviation {\small${\textrm{mean}}_{f}\min_{c}{\textrm{mean}}_{k\in Part_{c}}D_{f,k}$} to evaluate the location stability, where {\small$Part_{c}$} denotes the set of part indexes belonging to category $c$.

\subsubsection{Experimental results and analysis}

Tables~\ref{tab:interpretability} and \ref{tab:multi-interpretability} compare part interpretability of CNNs for single-category classification and that of CNNs for multi-category classification, respectively. Tables~\ref{tab:imgnet-stability}, \ref{tab:voc-stability}, and \ref{tab:cub200-stability} list average relative location deviations of CNNs for single-category classification. Table~\ref{tab:multi-stability} compares average relative location deviations of CNNs for multi-category classification. Our interpretable CNNs exhibited much higher interpretability and much better location stability than ordinary CNNs in almost all comparisons. Table~\ref{tab:classification} compares classification accuracy of different CNNs. Ordinary CNNs performed better in single-category classification. Whereas, for multi-category classification, interpretable CNNs exhibited superior performance to ordinary CNNs. The good performance in multi-category classification may be because that the clarification of filter semantics in early epochs reduced difficulties of filter learning in later epochs.

\begin{figure}[t]
\centering
\includegraphics[width=0.99\linewidth]{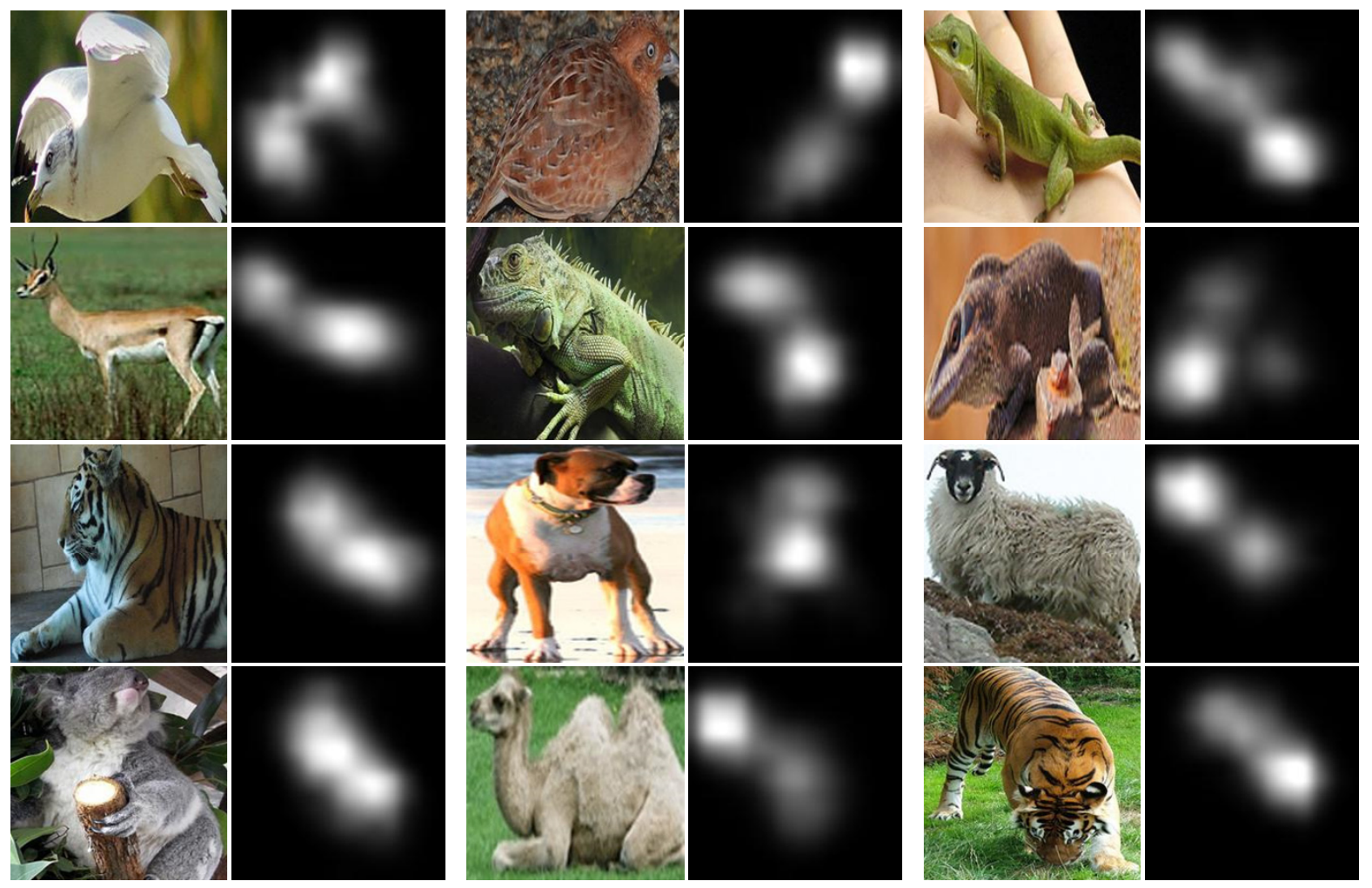}
\caption{Heat maps for distributions of object parts that are encoded in interpretable filters. We use all filters in the top conv-layer to compute the heat map.}
\label{fig:heatmap}
\vspace{-10pt}
\end{figure}

\subsection{Visualization of filters}

We followed the method proposed by Zhou~\emph{et al.}~\cite{CNNSemanticDeep} to compute the RF of neural activations of an interpretable filter, which was scaled up to the image resolution. Fig.~\ref{fig:visual} shows RFs\footnotemark[6] of filters in top conv-layers of CNNs, which were trained for single-category classification. Filters in interpretable CNNs were mainly activated by a certain object part, whereas filters in ordinary CNNs usually did not have explicit semantic meanings. Fig.~\ref{fig:heatmap} shows heat maps for distributions of object parts that were encoded in interpretable filters. Interpretable filters usually selectively modeled distinct object parts of a category and ignored other parts.

\begin{table}[t]
\centering
\resizebox{1.0\linewidth}{!}{\begin{tabular}{c|c|cc|ccc}
\hline
\!\!\!&\multicolumn{3}{|c}{multi-category} & \multicolumn{3}{|c}{single-category}\\
\cline{2-7}
\!\!\!\!\!&\!\!\! {\scriptsize ILSVRC Part} \!\!\!\!& \multicolumn{2}{|c|}{\scriptsize VOC Part} &\!\!\!\! {\scriptsize ILSVRC Part} \!\!\!\!\!\!\!&\!\!\!\!\!\! {\scriptsize VOC Part} \!\!\!\!\!\!\!&\!\!\!\!\!\! {\scriptsize CUB200}\!\!\\
\cline{2-4}
\!\!\!\!\!&\!\! {\small logistic\footnotemark[5]}\!\!\!\!\!\!\!&\!\!\!\!\!\! {\small logistic\footnotemark[5]}\!\!\!\!\!\!\!&\!\!\!\!\!\! {\small softmax} \!\!\!\!\!\!\!&\!\!\!\! \!\!\!\!\!\!\!&\!\!\!\!\!\! \!\!\!\!\!\!\!&\!\!\!\!\!\!\\
\hline
\!\!\!AlexNet \!\!\!\!\!&\!\! -- \!\!&\!\! -- \!\!&\!\! -- \!\!&\!\! {\bf96.28} \!\!&\!\! {\bf95.40} \!\!&\!\! {\bf95.59}
\\
\!\!\!{\footnotesize interpretable} \!\!\!\!\!&\!\! --\!\!&\!\! -- \!\!&\!\! -- \!\!&\!\! 95.38 \!\!&\!\! 93.93 \!\!&\!\! 95.35
\\
\hline
\!\!\!VGG-M \!\!\!\!\!&\!\! 96.73 \!\!&\!\! 93.88 \!\!&\!\! 81.93 \!\!&\!\! {\bf97.34} \!\!&\!\! {\bf96.82} \!\!&\!\! {\bf97.34}
\\
\!\!\!{\footnotesize interpretable} \!\!\!\!\!&\!\! {\bf97.99} \!\!&\!\! {\bf96.19} \!\!&\!\! {\bf88.03} \!\!&\!\! 95.77 \!\!&\!\! 94.17 \!\!&\!\! 96.03
\\
\hline
\!\!\!VGG-S \!\!\!\!\!&\!\! 96.98 \!\!&\!\! 94.05 \!\!&\!\! 78.15 \!\!&\!\! {\bf97.62} \!\!&\!\! {\bf97.74} \!\!&\!\! {\bf97.24}
\\
\!\!\!{\footnotesize interpretable} \!\!\!\!\!&\!\! {\bf98.72} \!\!&\!\! {\bf96.78} \!\!&\!\! {\bf86.13} \!\!&\!\! 95.64 \!\!&\!\! 95.47 \!\!&\!\! 95.82
\\
\hline
\!\!\!VGG-16 \!\!\!\!\!&\!\! -- \!\!&\!\! 97.97 \!\!&\!\! 89.71 \!\!&\!\! {\bf98.58} \!\!&\!\! {\bf98.66} \!\!&\!\! {\bf98.91}
\\
\!\!\!{\footnotesize interpretable} \!\!\!\!\!&\!\! -- \!\!&\!\! {\bf98.50} \!\!&\!\! {\bf91.60} \!\!&\!\! 96.67 \!\!&\!\! 95.39 \!\!&\!\! 96.51
\\
\hline
\end{tabular}}
\vspace{1pt}
\caption{Classification accuracy based on different datasets. In single-category classification, ordinary CNNs performed better, while in multi-category classification, interpretable CNNs exhibited superior performance.}
\label{tab:classification}
\vspace{-8pt}
\end{table}

\section{Conclusion and discussions}

In this paper, we have proposed a general method to modify traditional CNNs to enhance their interpretability. As discussed in \cite{Interpretability}, besides the discrimination power, the interpretability is another crucial property of a network. We design a loss to push a filter in high conv-layers toward the representation of an object part without additional annotations for supervision. Experiments have shown that our interpretable CNNs encoded more semantically meaningful knowledge in high conv-layers than traditional CNNs.

In future work, we will design new filters to describe discriminative textures of a category and new filters for object parts that are shared by multiple categories, in order to achieve a higher model flexibility.


{\small
\bibliographystyle{ieee}
\bibliography{TheBib}
}

\newpage
\onecolumn
\section*{Appendix}

\subsection*{Proof of equations}

\begin{equation}
\begin{split}
\frac{\partial{\bf Loss}}{\partial x_{ij}}=&-\sum_{T\in{\bf T}}p(T)\Big\{\frac{\partial p(x|T)}{\partial x_{ij}}\big[\log p(x|T)-\log p(x)+1\big]-p(x|T)\frac{\partial\log p(x)}{\partial x_{ij}}\Big\}\\
=&-\sum_{T\in{\bf T}}p(T)\Big\{\frac{\partial p(x|T)}{\partial x_{ij}}\big[\log p(x|T)-\log p(x)+1\big]-p(x|T)\frac{1}{p(x)}\frac{\partial p(x)}{\partial x_{ij}}\Big\}\\
=&-\sum_{T\in{\bf T}}p(T)\Big\{\frac{\partial p(x|T)}{\partial x_{ij}}\big[\log p(x|T)-\log p(x)+1\big]-p(x|T)\frac{1}{p(x)}\sum_{T'}\Big[p(T')\frac{\partial p(x|T')}{\partial x_{ij}}\Big]\Big\}\\
=&-\sum_{T\in{\bf T}}p(T)\Big\{\frac{\partial p(x|T)}{\partial x_{ij}}\big[\log p(x|T)-\log p(x)+1\big]\Big\}\\
&+\sum_{T\in{\bf T}}p(T)\frac{\partial p(x|T)}{\partial x_{ij}}\frac{\sum_{T'}p(T')p(x|T')}{p(x)}\qquad\qquad//\;\;\textrm{swap roles of $T$ and $T'$}\\
=&-\sum_{T\in{\bf T}}p(T)\Big\{\frac{\partial p(x|T)}{\partial x_{ij}}\big[\log p(x|T)-\log p(x)+1\big]\Big\}+\sum_{T\in{\bf T}}p(T)\frac{\partial p(x|T)}{\partial x_{ij}}\\
=&-\sum_{T\in{\bf T}}\frac{\partial p(x|T)}{\partial x_{ij}}p(T)\big[\log p(x|T)-\log p(x)\big]\\
=&-\sum_{T\in{\bf T}}\frac{t_{ij}p(T)e^{tr(x\cdot T)}}{Z_{T}}\Big\{tr(x\cdot T)-\log\big[Z_{T}p(x)\big]\Big\}
\end{split}
\nonumber
\end{equation}

\begin{equation}
\begin{split}
{\bf Loss}=&-MI({\bf X};{\bf T})\qquad\qquad//\;\;{\bf T}=\{T^{-},T_{\mu_{1}},T_{\mu_{2}},\ldots,T_{\mu_{n^2}}\}\\
=&-H({\bf T})+H({\bf T}|{\bf X})\\
=&-H({\bf T})-\sum_{x}p(x)\sum_{T\in{\bf T}}p(T|x)\log p(T|x)\\
=&-H({\bf T})-\sum_{x}p(x)\Big\{p(T^{-}|x)\log p(T^{-}|x)+\sum_{\mu}p(T_{\mu}|x)\log p(T_{\mu}|x)\Big\}\\
=&-H({\bf T})-\sum_{x}p(x)\Big\{p(T^{-}|x)\log p(T^{-}|x)+\sum_{\mu}p(T_{\mu}|x)\log\big[\frac{p(T_{\mu}|x)}{p({\bf T}^{+}|x)}p({\bf T}^{+}|x)\big]\Big\}\quad//p({\bf T}^{+}|x)=\sum_{\mu}p(T_{\mu}|x)\\
=&-H({\bf T})-\sum_{x}p(x)\Big\{p(T^{-}|x)\log p(T^{-}|x)+p({\bf T}^{+}|x)\log p({\bf T}^{+}|x)+\sum_{\mu}p(T_{\mu}|x)\log\frac{p(T_{\mu}|x)}{p({\bf T}^{+}|x)}\Big\}\\
=&-H({\bf T})+H({\bf T}'=\{T^{-},{\bf T}^{+}\}|{\bf X})+\sum_{x}p({\bf T}^{+},x)H({\bf T}^{+}=\{T_{\mu}\}|X=x)
\end{split}
\nonumber
\end{equation}
where
\begin{equation}
H({\bf T}''=\{T_{\mu}\}|X=x)=\sum_{\mu}\tilde{p}(T_{\mu}|X=x)\log\tilde{p}(T_{\mu}|X=x),\qquad\tilde{p}(T_{\mu}|X=x)=\frac{p(T_{\mu}|x)}{p({\bf T}^{+}|x)}\nonumber
\end{equation}

\subsection*{Visualization of CNN filters}

\begin{figure*}[h]
\centering
\includegraphics[width=0.75\linewidth]{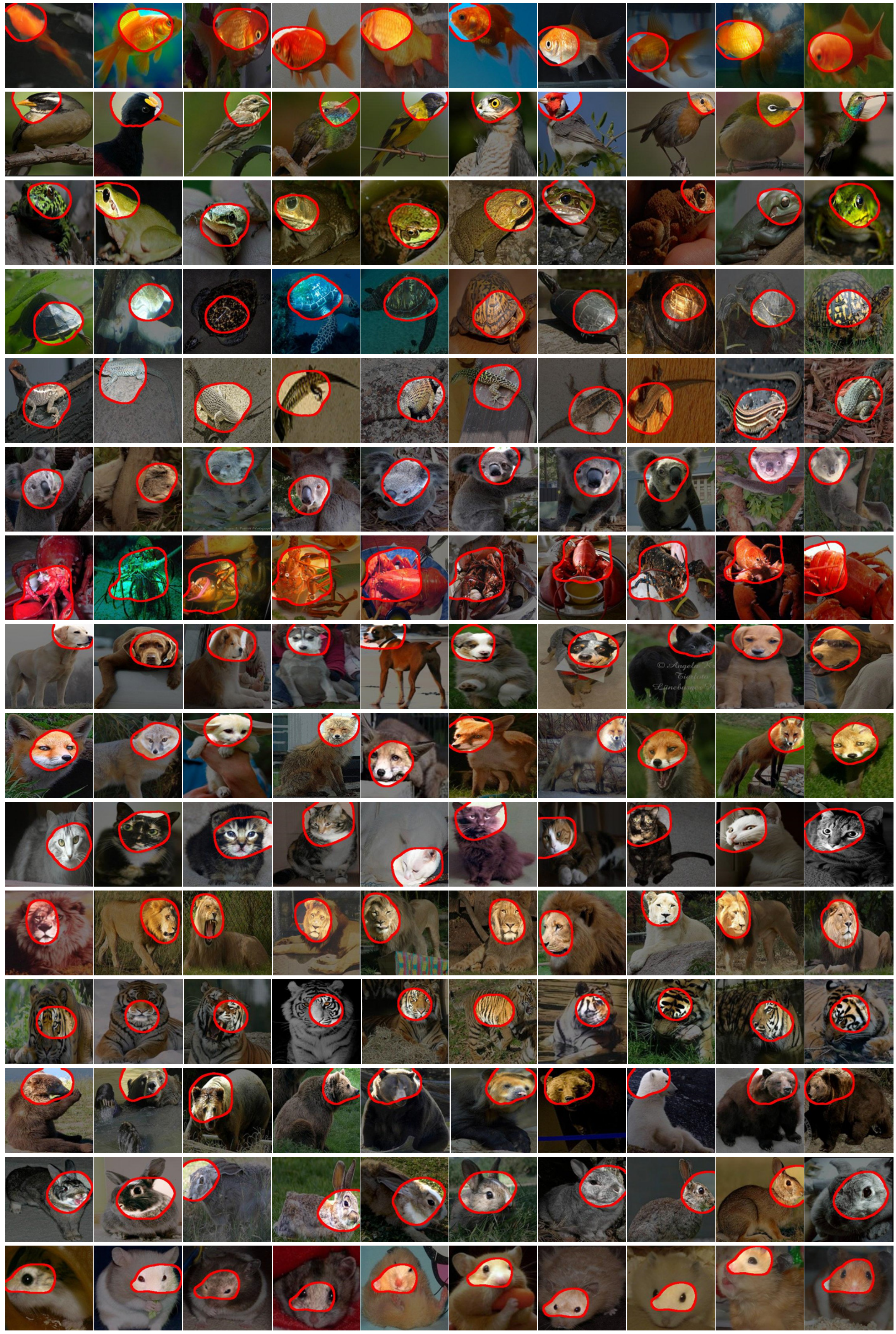}
\caption{Visualization of filters in the top interpretable conv-layer. Each row corresponds to feature maps of a filter in a CNN that is learned to classify a certain category.}
\end{figure*}

\begin{figure*}[h]
\centering
\includegraphics[width=0.75\linewidth]{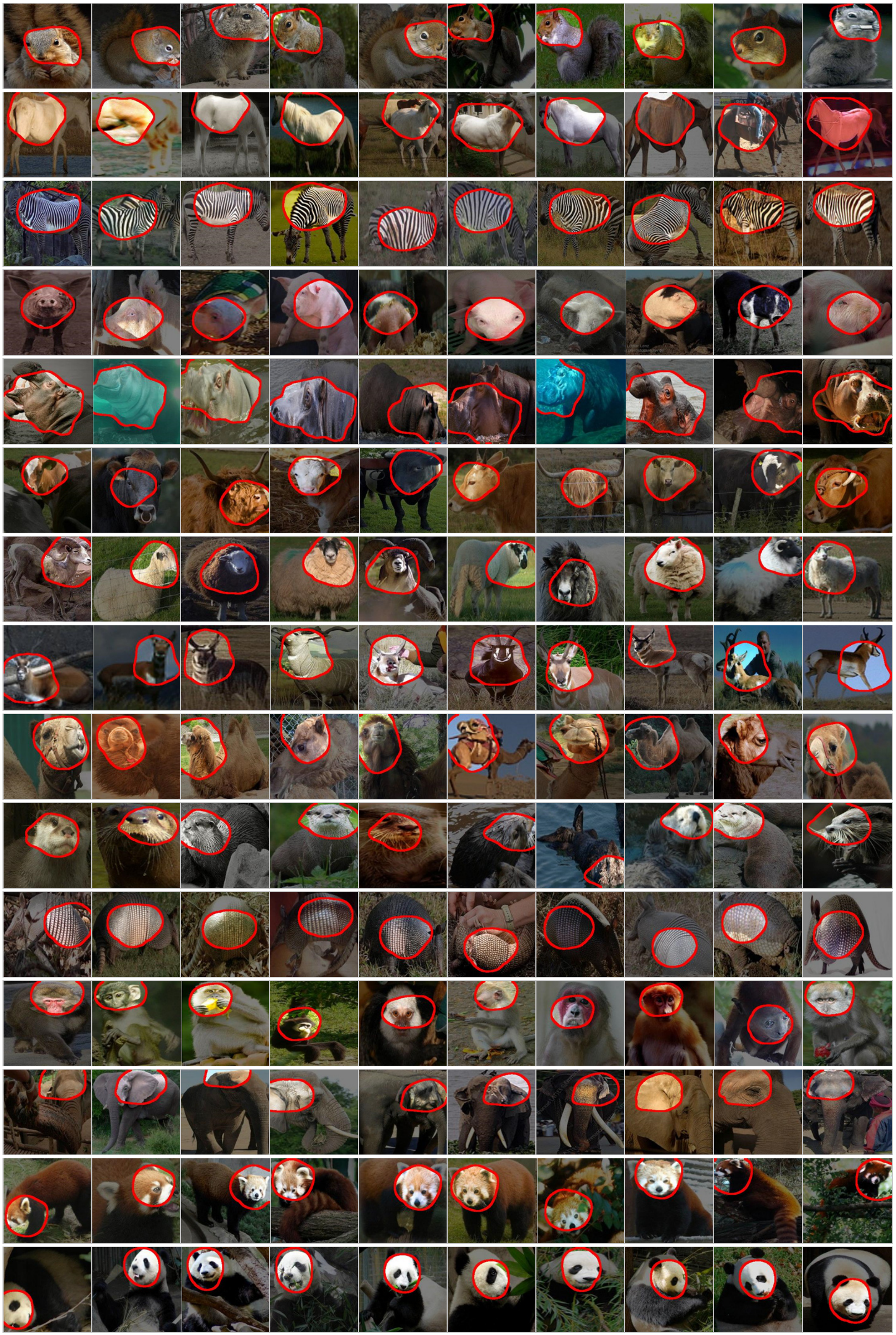}
\caption{Visualization of filters in the top interpretable conv-layer. Each row corresponds to feature maps of a filter in a CNN that is learned to classify a certain category.}
\end{figure*}

\begin{figure*}[h]
\centering
\includegraphics[width=0.75\linewidth]{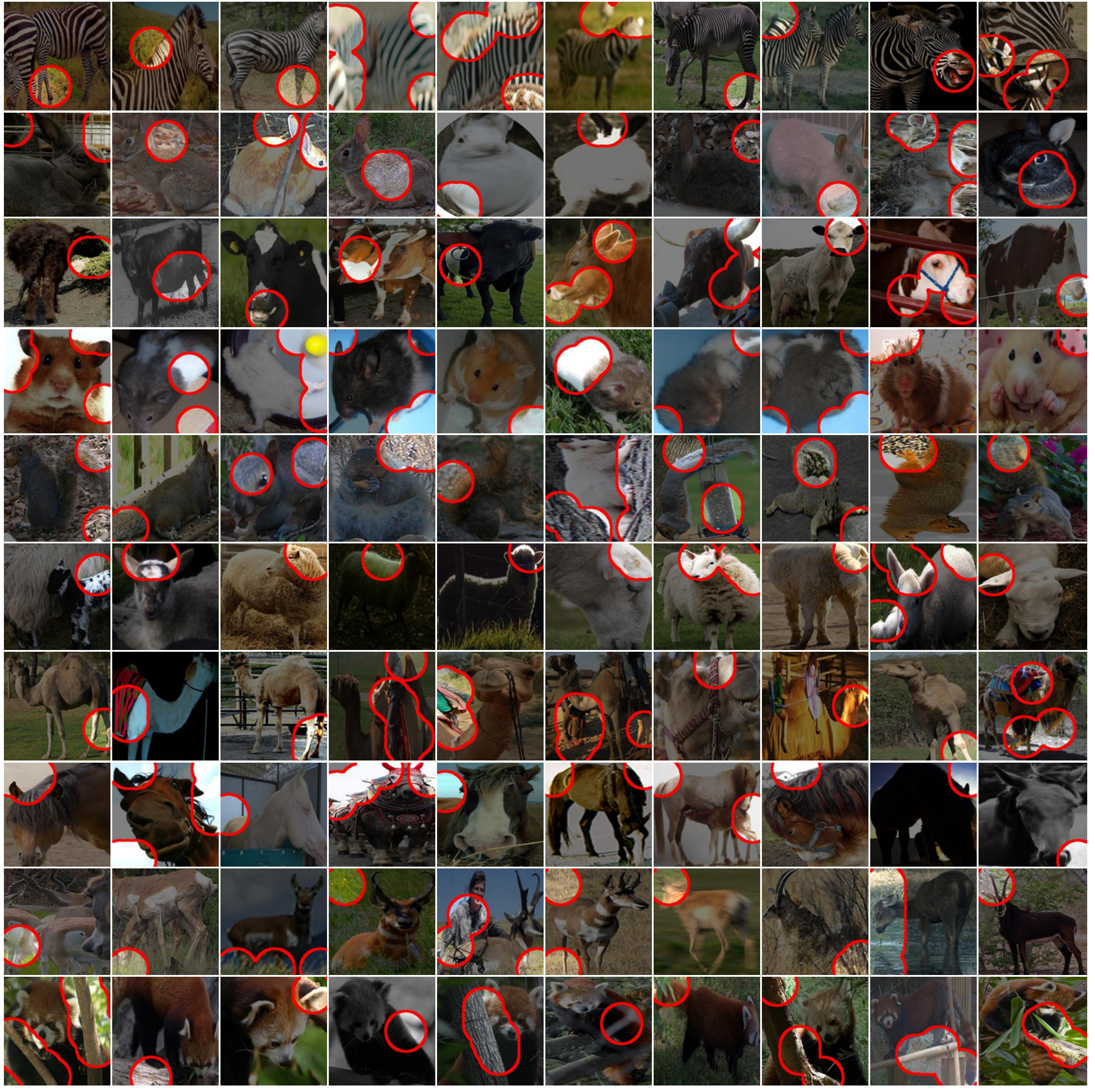}
\caption{Visualization of filters in the top conv-layer of an ordinary CNN. Each row corresponds to feature maps of a filter in a CNN that is learned to classify a certain category.}
\end{figure*}

\end{document}